\documentclass[review,12pt]{elsarticle}

\usepackage{amssymb}
\usepackage{amsmath}

\usepackage{xcolor}

\usepackage[hyphens]{url}   
\usepackage[colorlinks=true, allcolors=blue]{hyperref}
\journal{NeuroImage}

\begin{document}

\begin{frontmatter}



\title{Automated Detection of Multiple Sclerosis Lesions on 7-tesla MRI Using U-net and Transformer-based Segmentation}


\author[umd]{Michael Maynord} 
\ead{maynord@umd.edu}

\author[umd]{Minghui Liu}
\ead{minghui@umd.edu}

\author[umd]{Cornelia Ferm\"uller} 
\ead{fermulcm@umd.edu}

\author[umb]{Seongjin Choi}
\ead{Seongjin.Choi@som.umaryland.edu}

\author[umb]{Yuxin Zeng}
\ead{Yuxin.Zeng@som.umaryland.edu}

\author[umb]{Shishir Dahal}
\ead{SDahal@som.umaryland.edu}

\author[umb,vamc]{Daniel M. Harrison\corref{cor1}}
\ead{DHarrison@som.umaryland.edu}

\affiliation[umd]{organization={Department of Computer, Science University of Maryland College Park},
            city={College Park},
            postcode={20742}, 
            state={MD},
            country={USA}}

\cortext[cor1]{Corresponding author. Address: 110 S Paca Street, 3\textsuperscript{rd} Floor, Baltimore, MD 21201}
\affiliation[umb]{%
  organization={Department of Neurology, University of Maryland School of Medicine},
  city={Baltimore},
  state={MD},
  postcode={21201},
  country={USA}
}

\affiliation[vamc]{%
  organization={Baltimore VA Medical Center, VA Maryland Healthcare System},
  city={Baltimore},
  state={MD},
  postcode={21201},
  country={USA}
}

\begin{abstract}
Ultra-high field 7-tesla (7T) MRI improves visualization of multiple sclerosis (MS) white matter lesions (WML) but differs sufficiently in contrast and artifacts from 1.5–3T imaging - suggesting that widely used automated segmentation tools may not translate directly. We analyzed 7T FLAIR scans and generated reference WML masks from Lesion Segmentation Tool (LST) outputs followed by expert manual revision. As external comparators, we applied LST-LPA and the more recent LST-AI ensemble, both originally developed on lower-field data. We then trained 3D UNETR and SegFormer transformer-based models on 7T FLAIR at multiple resolutions (0.5$\times$0.5$\times$0.5~mm\textsuperscript{3}, 1.0$\times$1.0$\times$1.0~mm\textsuperscript{3}, and 1.5$\times$1.5$\times$2.0~mm\textsuperscript{3}) and evaluated all methods using voxel-wise and lesion-wise metrics from the BraTS 2023 framework. On the held-out test set at native 0.5$\times$0.5$\times$0.5~mm\textsuperscript{3} resolution, 7T-trained transformers achieved competitive overlap with LST-AI while recovering additional small lesions that were missed by classical methods, at the cost of some boundary variability and occasional artifact-related false positives. On a held-out 7 T test set, our best transformer model (SegFormer) achieved a voxel-wise Dice of 0.61 and lesion-wise Dice of 0.20, improving on the classical LST-LPA tool (Dice 0.39, lesion-wise Dice 0.02). Performance decreased for models trained on downsampled images, underscoring the value of native 7T resolution for small-lesion detection. By releasing our 7T-trained models, we aim to provide a reproducible, ready-to-use resource for automated lesion quantification in ultra-high field MS research\footnote{\url{https://github.com/maynord/7T-MS-lesion-segmentation}}. 
\end{abstract}



\begin{keyword}
7 Tesla \sep Multiple Sclerosis \sep Lesion Segmentation \sep Deep Learning \sep Computer Vision




\end{keyword}

\end{frontmatter}



\section{Introduction}

Multiple sclerosis (MS) is a chronic autoimmune and neurodegenerative disease that affects the central nervous system. It is the leading cause of non-traumatic neurological disability in young adults, with more than 2.8 million people affected worldwide \cite{GBD2019,Walton2020}. MS significantly reduces quality of life, causing mobility problems, fatigue, pain, and cognitive impairment. Life expectancy is shortened by approximately 5 to 10 years, largely due to secondary complications rather than the disease itself \cite{Marrie2015}. Given its high prevalence and long-term disability, improving diagnosis and disease monitoring in MS is a major clinical need.


Magnetic resonance imaging (MRI) is the gold standard for diagnosing MS and tracking its progression. MRI allows visualization of white matter lesions (WML), which are areas of inflammation and demyelination in the brain and spinal cord. Lesion burden and activity are core components of the McDonald criteria, the international diagnostic standard for MS, which were recently revised in 2024 to incorporate new MRI and biomarker evidence \cite{montalban2025diagnosis}. Early and accurate detection of lesions is essential, because treatment started soon after diagnosis can delay disease progression and reduce long-term disability \cite{Comi2017}.

Ultra-high field 7T MRI has emerged as a powerful tool for MS research. Compared with 3T MRI, 7T provides higher signal-to-noise ratio and finer spatial resolution, allowing visualization of smaller lesions and improved detection of cortical and juxtacortical pathology \cite{Absinta2018,Mainero2009}. Recent studies show that 7T can reveal lesions not visible at lower field strengths, supporting its growing use in clinical research \cite{deGraaf2022}. In 2017, the U.S. Food and Drug Administration (FDA) authorized the first 7T scanner for clinical use, accelerating its adoption in research centers \cite{FDA2017_press,FDA2017_K170840}.


Despite these advantages, 7T MRI also introduces unique challenges. \textbf{B0 inhomogeneity} refers to unevenness in the main magnetic field, which at 7T produces image distortions and signal voids, especially near the sinuses and temporal lobes \cite{Ugurbil2018}. \textbf{B1 inhomogeneity} arises from nonuniform distribution of the radiofrequency field, leading to intensity variations across the image, with central brightening and peripheral signal loss \cite{VanDeMoortele2005}. \textbf{Susceptibility artifacts}, caused by differences in magnetic properties of tissues, are exaggerated at 7T, creating dropout or blooming near air--tissue boundaries and veins \cite{Yacoub2003}. Finally, the \textbf{specific absorption rate (SAR)}, which measures RF energy deposited in tissue, increases at 7T, limiting sequence design and often requiring trade-offs in coverage or contrast \cite{Ladd2018}. These issues give 7T scans a distinctive appearance that differs substantially from 3T MRI, complicating both visual and automated lesion detection.


Automated lesion segmentation is essential for reproducibility and efficiency in MS research. However, most existing tools, such as Lesion Segmentation Tool (LST) implemented in Statistical and Parametric Mapping (SPM) software, were developed for 1.5T or 3T MRI \cite{Schmidt2012,Wiltgen2024}. LST provides classical, engineered lesion segmentation algorithms such as the Lesion Prediction Algorithm (LST-LPA) \cite{Schmidt2012}. LST-AI is a more recent open-source extension that uses an ensemble of three 3D U-Nets trained on 3T data \cite{Wiltgen2024}. In this work, we treat LST-LPA and LST-AI as established external baselines rather than architectures that we design or train.

In this study, we present the first publicly available models trained on 7T FLAIR MRI for MS lesion segmentation (\href{https://github.com/maynord/7T-MS-lesion-segmentation}{https://github.com/maynord/7T-MS-lesion-segmentation}). We systematically compare several deep learning architectures, including UNETR and SegFormer, with widely used conventional tools such as LST-LPA and LST-AI. We further examine how image resolution impacts performance. By releasing trained models, we provide a resource for researchers and clinicians who lack access to 7T training data but increasingly use 7T in their work. These contributions help close the gap in automated lesion segmentation for 7T MRI and support reproducible and standardized MS research.

\section{Methods}

\subsection{Approvals and consents}
Protocols were reviewed and approved by the institutional review boards at the University of Maryland, Baltimore and the Johns Hopkins University School of Medicine. All participants reviewed and signed written informed consent documents. 

\subsection{Study participants}
Imaging data for this study were derived from a previous longitudinal 7T MRI study that has been published in various forms elsewhere \cite{harrison2024meningeal, patel2023imaging, dahal2024pilot}. The data used here includes 287 scans from 138 study participants. This includes 103 participants with MS, 19 healthy controls (HC), and 16 with other inflammatory neurological disorders (OIND). Participants with MS and OIND were recruited from the University of Maryland Center for Multiple Sclerosis Treatment and Research. Participants with MS were required to meet 2017 revised McDonald Criteria \cite{thompson2018diagnosis} and included those with progressive forms of MS (PMS; including primary and secondary) and relapsing-remitting MS (RRMS). The mean (SD) age of MS participants was 45 (11) and 70 (68\%) were female. The OIND category included autoimmune/paraneoplastic encephalitis (7), idiopathic brainstem inflammatory lesion (1), chronic relapsing inflammatory optic neuropathy (1), neuroborreliosis (1), neuromyelitis optica (4), Sjogren myelopathy (1), transverse myelitis (1), and inflammatory vasculitis (1). The mean (SD) age of OIND participants was 43 (16) and 11 (69\%) were female. HC were recruited by advertisements and were approximately matched to MS participants for sex and age. The mean (SD) age of HC participants was 42 (13) and 12 (63\%) were female. The data reported here includes recruitment that occurred from October 2019 through December 2023.

\subsection{MRI acquisition}
MRI of the brain was acquired on a 7T Phillips Achieva scanner (Phillips Healthcare, Best, The Netherlands) with an 8-channel, multi-transmit and 32 channel receive head coil (NovaMedical, Wilmington, MA USA). A Magnetization Prepared 2 Rapid Gradient Echo (MP2RAGE) sequence was acquired with the following parameters: MP2RAGE TR = 8250 ms, TR = 6.9 ms, TE= 1.97 ms, inversion times = 1000/3300 ms, flip angles = 5/5 °, Turbo factor = 252, Field of view = 220 × 220 mm2, near isotropic resolution = 0.700 × 0.688 × 0.688 mm3, SENSE acceleration factor = 2 × 2 and acquisition time = 9 min 36 s. A three dimensional (3D) Fluid Attenuated Inversion Recovery (FLAIR) sequence was also acquired with the following parameters: TR = 8000 ms, TE = 300 ms, TI = 2200 ms, flip angle = 70-degree, near isotropic resolution = 0.500 × 0.488 × 0.488 mm3, SENSE acceleration factor = 2.5 × 3.0 and acquisition time = 9 min 4 s.

\subsection{Data Processing}

\subsubsection{Image processing}

7T MRI images were processed to generate MP2RAGE T1-weighted (T1W) images using custom code \cite{marques2010mp2rage}.  After N4 inhomogeneity correction \cite{tustison2010n4itk},  these T1W images were denoised by multiplying the image by the second inversion time image in the corresponding MP2RAGE acquisition. FLAIR images were co-registered to the denoised pre-contrast T1W images following N4 inhomogeneity correction utilizing ANTs (Advanced Normalization Tools, version 1.10.14) \cite{avants2011reproducible, tustison2021antsx}.



\subsubsection{Initial Segmentation}

Due to the extreme high resolution of these images, pure WML segmentation ‘from scratch’ on the entire dataset was not feasible in terms of person-hours available. Thus, we took an approach of initial segmentation with LST with manual revision to create our manual ‘gold standard’ masks. WMLs were initially segmented on FLAIR images from each scanning session using either the LST-LPA - the Lesion Prediction Algorithm (LPA) from the Lesion Segmentation Tool (LST, version 3.0.0) \cite{schmidt2012automated} for the Statistical Parametric Mapping (SPM, version 12) or the LST-AI algorithm \cite{wiltgen2024lst}. Outputs from LST were then manually reviewed by expert and trained raters (DMH, YZ, SD, SC) in ITK-SNAP software \cite{yushkevich2006user}. Image analysis tools in ITK-SNAP were utilized to revise the WML masks, removing false-positive voxels and filling in regions that were missed by the segmentation algorithm. The final, manually revised mask was saved and was utilized going forward as the ‘gold standard’ WML mask for further assessment and training tasks.

For scans assigned to the training and validation sets, only the LST-LPA probability maps were used as algorithmic pre-labels prior to manual editing. For the held-out test set, raters were additionally shown LST-AI segmentations and could optionally use them as starting masks during editing; LST-AI outputs were not used in constructing any training labels and were never seen by the models during training.

In addition to the ‘gold standard’, manually revised masks, we also reserved the original LST-LPA outputs (with the 0.3 probability threshold) and the results of another segmentation using the newer LST-AI algorithm \cite{wiltgen2024lst} as comparators to our segmentation results.


\subsection{Model architectures}

We evaluated four segmentation methods for white matter lesion detection on 7T FLAIR: two established tools from the Lesion Segmentation Tool (LST) framework, LST-LPA and LST-AI, and two deep learning models, UNETR and SegFormer, trained directly on 7T data.

\subsubsection{LST-LPA}

LST-LPA is a classical probabilistic method implemented in the Lesion Segmentation Tool (LST, version 3.0.0) for Statistical Parametric Mapping (SPM, version 12) \cite{schmidt2012automated}. It models lesion probability using logistic regression on hand-crafted intensity and spatial features derived primarily from FLAIR images and outputs a continuous lesion probability map, which we threshold at 0.3 to obtain a binary lesion mask.

\subsubsection{LST-AI}

LST-AI is a deep learning extension of the LST framework that uses an ensemble of three 3D U-Net–style convolutional networks trained on 3T MRI data \cite{wiltgen2024lst}. For this study, we applied the released LST-AI model in its default configuration to the 7T FLAIR and T1-weighted images and used the resulting probability maps set at a threshold of $0.5$ as one of the external comparator segmentations.

\subsubsection{UNETR}

UNETR is a hybrid transformer–U-Net architecture for volumetric segmentation \cite{Hatamizadeh2022}. The input volume is partitioned into non-overlapping 3D patches that are embedded and processed by a transformer encoder; intermediate encoder representations are then passed via skip connections to a U-Net–style decoder that performs multi-scale upsampling and produces voxelwise lesion probabilities. In our implementation, UNETR operates on 7T FLAIR images and outputs a single-channel probability map in the native 0.5$\times$0.5$\times$0.5~mm\textsuperscript{3} space, which is thresholded to obtain binary lesion masks.

\subsubsection{SegFormer}

SegFormer uses a hierarchical transformer encoder (MiT) with a lightweight multi-layer perceptron decoder for semantic segmentation \cite{Xie2021}. The encoder produces feature maps at multiple resolutions, which are projected to a common dimension and fused in the decoder to generate dense predictions. We adapt SegFormer to 3D 7T FLAIR inputs by applying the encoder–decoder to volumetric patches and producing voxelwise lesion probability maps in the native 0.5$\times$0.5$\times$0.5~mm\textsuperscript{3} resolution.

\subsubsection{Implementation and training}

We implemented UNETR and SegFormer in PyTorch and trained both models from scratch on the 7T FLAIR dataset. Scans were divided at the subject level into training and test sets (181/77 scans, respectively), ensuring that no participant appeared in both splits. We did not use ImageNet or other natural-image pretraining, because the single-channel 7T FLAIR contrast differs substantially from RGB images and our objective was to learn task-specific features from high-field data. We also did not employ pretraining on alternate volumetric datasets; each 7T scan provides full-brain, high-resolution coverage with numerous lesions, yielding a large number of supervised lesion and non-lesion voxels. Both models were trained using standard 3D data augmentation (e.g., intensity and spatial perturbations) and regularization to reduce overfitting. All four methods output voxelwise lesion probability maps that are thresholded to binary masks, enabling quantitative comparison under the common evaluation framework described in Section~\ref{sec:evaluation-metrics}.

\subsection{Evaluation Metrics}
\label{sec:evaluation-metrics}
We adopt the evaluation framework used in the BraTS 2023 Lesion-wise Performance Metrics challenge~\cite{BraTS2023Metrics}, which emphasizes both voxel-level and lesion-level agreement between predicted and ground-truth segmentations. This framework provides a comprehensive assessment of segmentation accuracy, overlap, and boundary precision.

\paragraph{Basic lesion-level and voxel-level counts.}
For lesion-wise metrics, $TP$, $FP$, and $FN$ denote the numbers of true-positive, false-positive, 
and false-negative \emph{lesions}, determined by lesion overlap. 
Lesion-level true negatives are not defined, as the background space is continuous. 
In contrast, voxel-wise metrics such as sensitivity and specificity are computed from 
the voxel-level confusion matrix, where $TP_v$, $FP_v$, $FN_v$, and $TN_v$ denote the 
numbers of correctly and incorrectly classified voxels:
\begin{align}
\text{Sensitivity} &= \frac{TP_v}{TP_v + FN_v} \label{eq:sensitivity}\\[2pt]
\text{Specificity} &= \frac{TN_v}{TN_v + FP_v} \label{eq:specificity}
\end{align}

\paragraph{Voxel-wise (legacy) overlap and boundary.}
Let $P$ and $G$ be the predicted and ground-truth voxel sets (binary masks).
The (legacy) Dice overlap is
\begin{equation}
\text{Dice}_{\text{legacy}} = \frac{2\,|P \cap G|}{|P| + |G|} \label{eq:dice-legacy}
\end{equation}
For boundaries $A=\partial P$ and $B=\partial G$, the Hausdorff distance (HD) and its 95th percentile variant (HD95) are
\begin{align}
HD(A,B) &= \max\Big\{ \sup_{a\in A}\inf_{b\in B} d(a,b),\ \sup_{b\in B}\inf_{a\in A} d(a,b) \Big\} \label{eq:hd}\\[2pt]
HD_{95}(A,B) &= \text{percentile}_{95}\Big(\{d(a,B)\mid a\in A\} \cup \{d(b,A)\mid b\in B\}\Big) \label{eq:hd95}
\end{align}
with $d(\cdot,\cdot)$ the Euclidean distance. HD95 downweights extreme outliers, reflecting boundary alignment for the bulk (95\%) of boundary points. This is particularly useful in medical imaging, where small boundary misregistrations may not be clinically meaningful yet can heavily influence overlap scores.

\paragraph{Lesion-wise metrics (BraTS 2023).}
Voxel-level metrics overweight large lesions. Lesion-wise metrics instead compute a per-lesion score and then average \emph{over lesions}, balancing small and large foci. Following BraTS 2023, let $L$ be the number of ground-truth lesions, and for each matched lesion $l_i$ compute $\text{Dice}(l_i)$ and $HD_{95}(l_i)$. Then
\begin{align}
\text{Lesion-wise Dice} &= 
\frac{\sum_{i=1}^{L} \text{Dice}(l_i)}{TP + FN + FP} \label{eq:lw-dice}\\[2pt]
\text{Lesion-wise }HD_{95} &=
\frac{\sum_{i=1}^{L} HD_{95}(l_i)}{TP + FN + FP} \label{eq:lw-hd95}
\end{align}
These definitions penalize both missed (\(FN\)) and spurious (\(FP\)) lesions directly in the denominator, so each lesion (or miss/spurious detection) contributes equally.

\paragraph{Preprocessing for lesion-wise evaluation.}
In accordance with BraTS 2023~\cite{BraTS2023Metrics}, lesion-wise evaluation is computed after two steps: (i) \emph{morphological dilation} of both prediction and ground-truth masks by 2 voxels, and (ii) a \emph{minimum lesion size threshold} of 5 voxels. These steps are applied \emph{only} to lesion-level evaluation (the $TP/FP/FN$ counts and the lesion-wise metrics in Eqs.~\ref{eq:lw-dice}--\ref{eq:lw-hd95}); they are \emph{not} applied to voxel-wise metrics (Eqs.~\ref{eq:dice-legacy}--\ref{eq:hd95}) nor to sensitivity/specificity (Eqs.~\ref{eq:sensitivity}--\ref{eq:specificity}). Dilation tolerates small boundary mismatches; the size threshold filters out tiny, likely spurious foci.

\paragraph{Evaluation resolution.}
All models were evaluated in the native 7T image space at 
0.5$\times$0.5$\times$0.5~mm\textsuperscript{3} resolution. 
Outputs from models trained at lower resolutions 
(1.0$\times$1.0$\times$1.0~mm\textsuperscript{3} and 1.5$\times$1.5$\times$2.0~mm\textsuperscript{3}) 
were upsampled to this common grid before metric computation to provide a uniform basis for comparison. Using the native 7T resolution ensures that all models are assessed in the imaging space that captures the full spatial detail available at ultra-high field strength. This enables a fair evaluation across architectures while directly testing the benefit of models trained on high-resolution 7T inputs relative to those trained on coarser data.

\section{Results}


\subsection{Quantitative results}


Across the set of quantitative metrics (Tables~\ref{tab:detection_summary}--\ref{tab:boundary_metrics}), all methods achieved nonzero lesion detection performance on 7T FLAIR, but with clear differences between classical and deep learning approaches. At the native 0.5$\times$0.5$\times$0.5~mm\textsuperscript{3} resolution, LST-AI obtained the highest voxel-wise Dice (0.68), the best lesion-wise Dice (0.45), and the lowest HD95 and lesion-wise HD95 (15.1 and 134.7, respectively), while maintaining relatively low false-positive counts (14.6 FP on average) and high specificity ($\geq 0.9997$). LST-LPA showed the highest voxel-wise sensitivity (0.82) but at the cost of substantially more false positives (748.2 FP) and lower Dice (0.39) and lesion-wise Dice (0.02). Among the transformer-based models, SegFormer slightly outperformed UNETR at 0.5$\times$0.5$\times$0.5~mm\textsuperscript{3} in terms of Dice (0.61 vs.\ 0.57) and HD95 (18.5 vs.\ 49.4), with both achieving similar sensitivities ($\sim$0.71) and specificities ($\geq 0.9996$). For UNETR and SegFormer, training at coarser resolutions (1.0$\times$1.0$\times$1.0~mm\textsuperscript{3} and 1.5~$\times$~1.5~$\times$~2~mm\textsuperscript{3}) led to a progressive decrease in Dice and lesion-wise Dice, higher lesion-wise HD95 (including an infinite value for SegFormer at the coarsest resolution), and increased false negatives, indicating that performance degrades as resolution is reduced. Overall, these results suggest that all methods benefit from the native high-resolution 7T data, with LST-AI providing the strongest numerical baseline; however, because LST-AI contributed to the generation of some of the reference segmentations, its scores should be interpreted as an optimistic upper bound rather than a fully independent comparator (see Discussion).

\begin{table}[h!]
\centering
\begin{tabular}{|l|c|c|c|c|c|c|}
\hline
\textbf{Model} & \textbf{Resolution} & \textbf{TP} & \textbf{FP} & \textbf{FN} & \textbf{Sensitivity} & \textbf{Specificity} \\
\hline
LST-LPA   & 0.5$\times$0.5$\times$0.5~mm\textsuperscript{3}           & 25.2 & 748.2 & 8.0  & 0.8211 & 0.9978 \\
LST-AI    & 0.5$\times$0.5$\times$0.5~mm\textsuperscript{3}           & 29.3 & 14.6  & 3.9  & 0.6980 & 0.9997 \\
UNETR     & 0.5$\times$0.5$\times$0.5~mm\textsuperscript{3}           & 29.1 & 128.5 & 4.1  & 0.7089 & 0.9996 \\
UNETR     & 1.0$\times$1.0$\times$1.0~mm\textsuperscript{3}           & 28.4 & 37.7  & 4.8  & 0.5953 & 0.9995 \\
UNETR     & 1.5$\times$1.5$\times$2.0 mm\textsuperscript{3} & 20.0 & 65.3  & 13.2 & 0.3665 & 0.9996 \\
SegFormer & 0.5$\times$0.5$\times$0.5~mm\textsuperscript{3}           & 27.0 & 42.2  & 4.6  & 0.7072 & 0.9997 \\
SegFormer & 1.0$\times$1.0$\times$1.0~mm\textsuperscript{3}           & 19.1 & 9.4   & 12.5 & 0.5500 & 0.9996 \\
SegFormer & 1.5$\times$1.5$\times$2.0 mm\textsuperscript{3} & 6.0  & 1.7   & 25.6 & 0.2362 & 0.9998 \\
\hline
\end{tabular}
\caption{Detection summary across models and resolutions: mean True Positives (TP), False Positives (FP), False Negatives (FN), sensitivity, and specificity on the held-out 7T test set.}
\label{tab:detection_summary}
\end{table}

\begin{table}[h!]
\centering
\begin{tabular}{|l|c|c|c|}
\hline
\textbf{Model} & \textbf{Resolution} & \textbf{DICE} & \textbf{Lesion-wise DICE} \\
\hline
LST-LPA   & 0.5$\times$0.5$\times$0.5~mm\textsuperscript{3}           & 0.3895 & 0.0184 \\
LST-AI    & 0.5$\times$0.5$\times$0.5~mm\textsuperscript{3}           & 0.6829 & 0.4495 \\
UNETR     & 0.5$\times$0.5$\times$0.5~mm\textsuperscript{3}           & 0.5657 & 0.0997 \\
UNETR     & 1.0$\times$1.0$\times$1.0~mm\textsuperscript{3}           & 0.5180 & 0.1534 \\
UNETR     & 1.5$\times$1.5$\times$2.0 mm\textsuperscript{3} & 0.3462 & 0.0712 \\
SegFormer & 0.5$\times$0.5$\times$0.5~mm\textsuperscript{3}           & 0.6120 & 0.1967 \\
SegFormer & 1.0$\times$1.0$\times$1.0~mm\textsuperscript{3}           & 0.5023 & 0.1761 \\
SegFormer & 1.5$\times$1.5$\times$2.0 mm\textsuperscript{3} & 0.2879 & 0.0656 \\
\hline
\end{tabular}
\caption{Overlap metrics across models and resolutions: voxel-wise Dice-Sørensen coefficient (DICE) and lesion-wise DICE computed in the native 0.5~mm\textsuperscript{3} 7T space.}
\label{tab:overlap_metrics}
\end{table}

\begin{table}[h!]
\centering
\begin{tabular}{|l|c|c|c|}
\hline
\textbf{Model} & \textbf{Resolution} & \textbf{HD95} & \textbf{Lesion-wise HD95} \\
\hline
LST-LPA   & 0.5$\times$0.5$\times$0.5~mm\textsuperscript{3}           & 21.08  & 359.98 \\
LST-AI    & 0.5$\times$0.5$\times$0.5~mm\textsuperscript{3}           & 15.13  & 134.71 \\
UNETR     & 0.5$\times$0.5$\times$0.5~mm\textsuperscript{3}           & 49.42  & 309.40 \\
UNETR     & 1.0$\times$1.0$\times$1.0~mm\textsuperscript{3}           & 22.19  & 238.66 \\
UNETR     & 1.5$\times$1.5$\times$2.0 mm\textsuperscript{3} & 48.78  & 302.21 \\
SegFormer & 0.5$\times$0.5$\times$0.5~mm\textsuperscript{3}           & 18.48  & 248.73 \\
SegFormer & 1.0$\times$1.0$\times$1.0~mm\textsuperscript{3}           & 16.25  & 207.39 \\
SegFormer & 1.5$\times$1.5$\times$2.0 mm\textsuperscript{3} & $\infty$ & 300.18 \\
\hline
\end{tabular}
\caption{Boundary metrics across models and resolutions: voxel-wise 95th percentile Hausdorff Distance (HD95) and lesion-wise HD95 in the native 0.5~mm\textsuperscript{3} 7T space.}
\label{tab:boundary_metrics}
\end{table}


\subsection{Qualitative results}

\begin{figure}[htbp]
    \centering
    \includegraphics[width=0.95\textwidth]{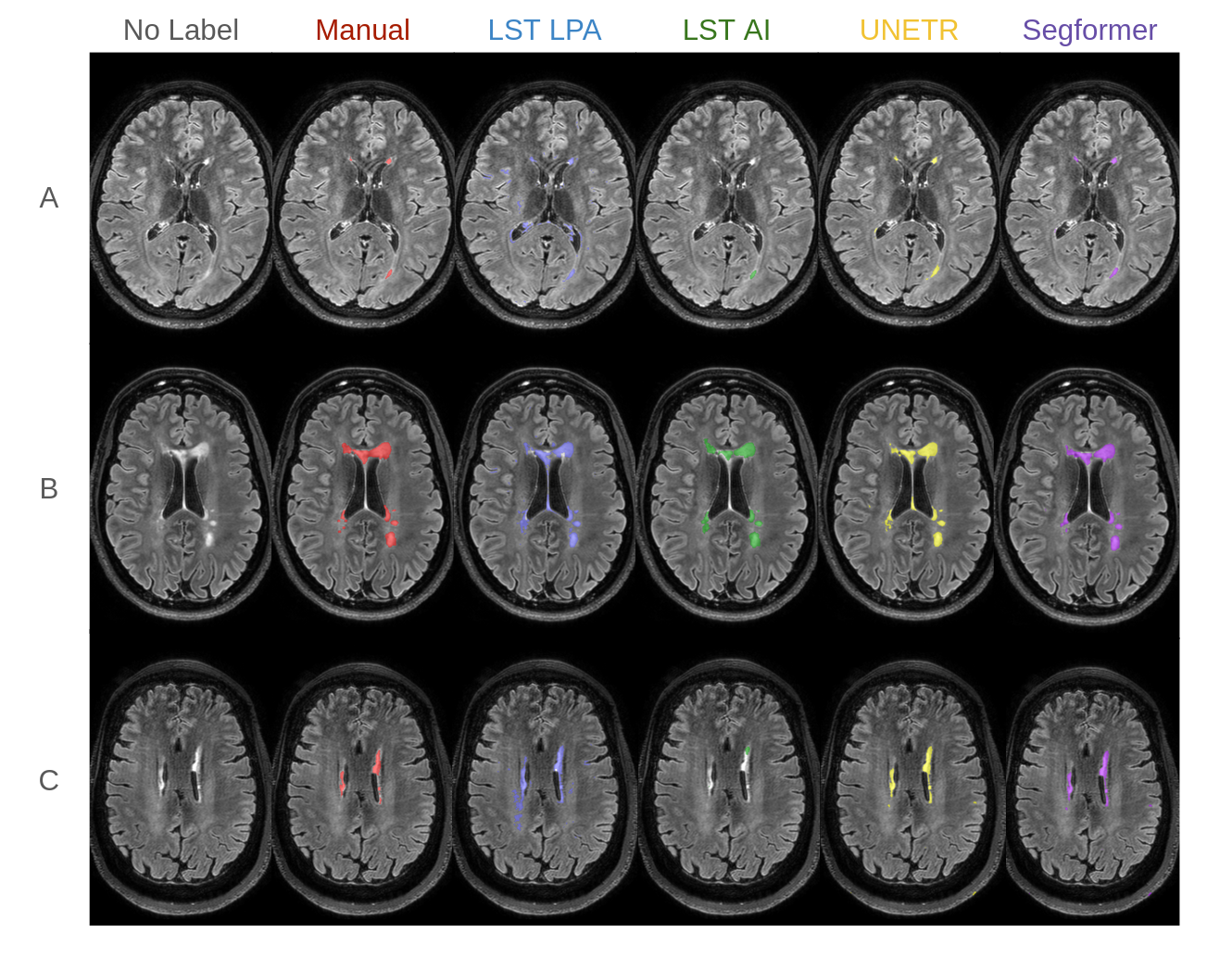}
    \caption{Comparison of segmentation methods. Shown are 7T FLAIR 0.5$\times$0.5$\times$0.5~mm\textsuperscript{3} images from three participants (A, B, and C) in this study and corresponding segmentations performed and/or trained on the 0.5$\times$0.5$\times$0.5~mm\textsuperscript{3} images. Visual inspection highlights qualitative differences between segmentation outputs from each method. Manual (red) masks indicate the manually annotated image by an expert rater, which acted as the ‘gold standard’ to which quantitative comparisons were performed. LST LPA appears to erroneously highlight bright cortical boundaries, which are often prominent on 7T FLAIR, as lesion, in addition to choroid plexus (A) and septum pellucidum (B). LST AI avoids many of the errors performed by LST LPA, but misses some lesions (left frontal in A) and underfills others (left frontal in C). UNETR and Segformer performed relatively well compared to manual annotation, in addition to capturing small lesions missed by the manual rater (small right subcortical WML in B and C). However, both occasionally erroneously highlighted false positives due to bright cortical boundaries (left frontoparietal in C). }
    \label{fig:qualitative_main}
\end{figure}

\begin{figure}[htbp]
    \centering
    \includegraphics[width=1.0\textwidth]{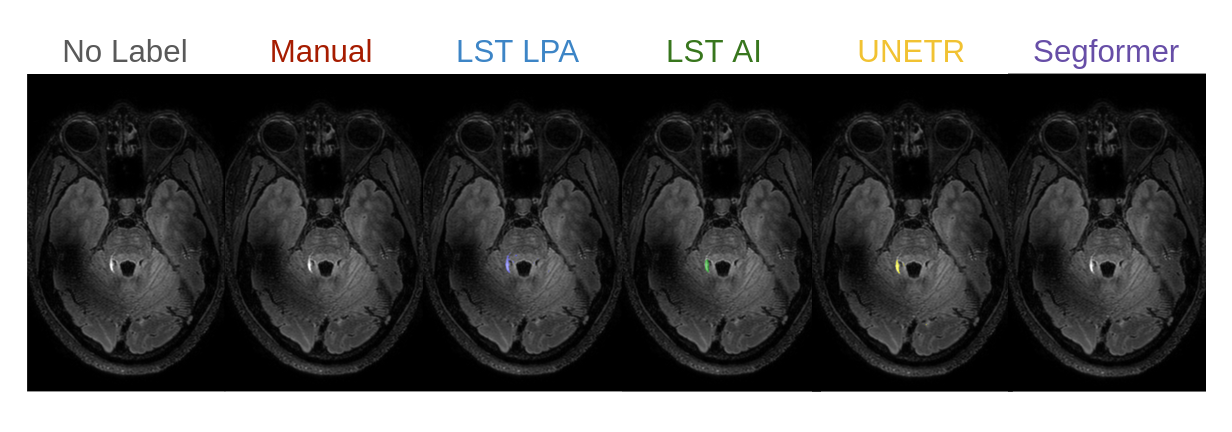}
    \caption{Handling of image artifacts. Shown are 7T FLAIR images (0.5$\times$0.5$\times$0.5~mm\textsuperscript{3}) from one participant in this study in which a wraparound artifact, likely from ear tissue, occurs in the region of the pons and cerebellum. Manual annotation did not identify this as lesion, but LST LPA, LST AI, and UNETR masked this as a false positive lesion. Segformer did not.}
    \label{fig:artifact}
\end{figure}

\begin{figure}[htbp]
    \centering
    \includegraphics[width=0.8\textwidth]{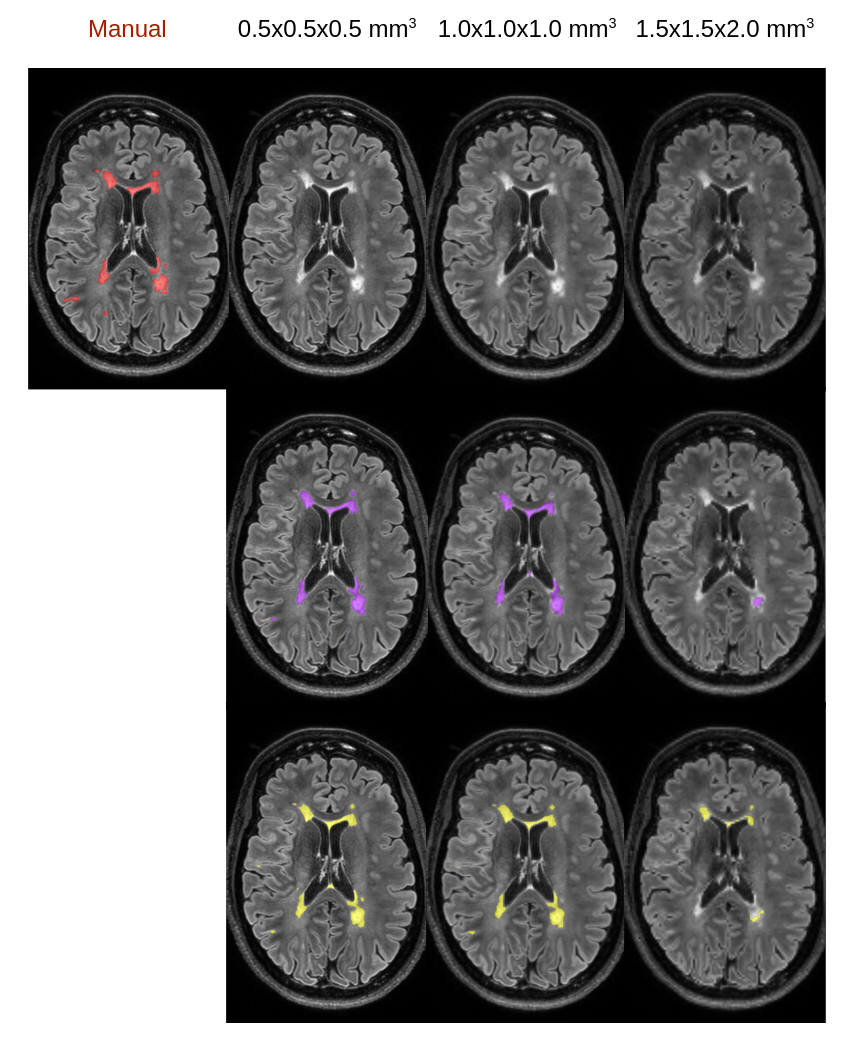}
    \caption{Comparison of models trained at various image resolutions. Shown are 7T FLAIR images from a participant with MS at the original 0.5$\times$0.5$\times$0.5~mm\textsuperscript{3} resolution and downsampled to 1.0$\times$1.0$\times$1.0~mm\textsuperscript{3} and 1.5$\times$1.5$\times$2.0~mm\textsuperscript{3} resolution (first row), along with a manually drawn lesion mask (red). UNETR segmentation shown in yellow and Segformer segmentation shown in purple. The model trained on images and masks in the 0.5$\times$0.5$\times$0.5~mm\textsuperscript{3} resolution space appear closest in segmentation accuracy to the manual, ‘ground-truth’ masks, with a slight reduction in accuracy seen for downsampled 1.0$\times$1.0$\times$1.0~mm\textsuperscript{3}. Downsampling to 1.5$\times$1.5$\times$2.0~mm\textsuperscript{3} clearly demonstrates segmentation inferiority to either isotropic option.}
    \label{fig:resolution}
\end{figure}

Qualitative examples illustrate how the different methods behave on individual 7T FLAIR scans (Figures~\ref{fig:qualitative_main}--\ref{fig:resolution}). As seen in the examples provided in Figure~\ref{fig:qualitative_main}, we frequently found that LST-LPA oversegmented bright cortical boundaries, choroid plexus, and septum pellucidum as lesions, whereas LST-AI substantially reduced these false positives but misses some lesions and underfilled others relative to the manual reference masks. UNETR and SegFormer generally produced segmentations that more closely followed the expert annotations and, in some cases, recovered small WML not captured in the original manual masks, while still occasionally labeling bright cortical boundaries as lesions. Figure~\ref{fig:artifact} shows a wraparound artifact near the pons and cerebellum: the manual annotation correctly excluded this region, whereas LST-LPA, LST-AI, and UNETR labeled it as lesion, and only SegFormer largely avoided this false-positive focus. Finally, in the resolution comparison in Figure~\ref{fig:resolution}, models trained at the native 0.5$\times$0.5$\times$0.5~mm\textsuperscript{3} resolution produced masks that most closely matched the manual reference, with visibly reduced lesion coverage and boundary accuracy at 1.0$\times$1.0$\times$1.0~mm\textsuperscript{3} and especially at 1.5$\times$1.5$\times$2.0~mm\textsuperscript{3}. These qualitative patterns align with the quantitative trends in the tables, highlighting both the strengths and remaining failure modes of each method on high-field 7T data.

\section{Discussion}



In this work, we evaluated classical and deep learning methods for automated WML segmentation on high-resolution 7T FLAIR MRI in Figures ~\ref{fig:qualitative_main}-\ref{fig:resolution}. The four methods we studied embody distinct inductive biases for lesion detection. LST-LPA combines hand-crafted intensity and spatial features with a logistic regression model, making strong assumptions about the appearance of lesions and their relationship to surrounding tissue. LST-AI replaces hand-crafted features with a U-Net–style convolutional encoder–decoder trained on large 3T datasets, learning local, multi-scale features directly from data. UNETR and SegFormer extend these deep-learning strategies with transformer encoders that explicitly model long-range dependencies and multi-scale context via self-attention. Conceptually, this progression from hand-crafted features to convolutional hierarchies to transformers corresponds to a shift from local, intensity-based decision rules toward architectures that can integrate fine-grained spatial detail with distributed contextual information—properties that are particularly relevant for small, scattered lesions and complex artifact patterns at 7T.

Classical and deep learning methods showed complementary strengths. At 0.5$\times$0.5$\times$0.5~mm\textsuperscript{3}, LST-LPA achieved the highest voxel-wise sensitivity but at the cost of substantial over-segmentation, reflected in a large number of false positives and lower Dice and lesion-wise Dice scores. LST-AI provided a more favorable balance of sensitivity, specificity, overlap, and boundary accuracy, achieving the highest Dice and lesion-wise Dice and the lowest HD95 and lesion-wise HD95 among all methods. The transformer-based models trained on 7T data approached but did not surpass LST-AI’s overlap-based metrics, with SegFormer slightly outperforming UNETR in Dice and HD95 at native resolution, while both maintained sensitivities comparable to LST-AI and much higher than those observed at coarser resolutions. These findings indicate that deep learning methods clearly outperform purely classical intensity-based tools such as LST-LPA on 7T FLAIR, and that domain-adapted transformer architectures can achieve performance that is competitive with, though not uniformly superior to, a strong LST-AI baseline.


Deep learning has rapidly advanced medical image segmentation by learning complex, hierarchical features directly from imaging data. In MS, convolutional neural networks such as U-Net and its derivatives have become the dominant approach, consistently outperforming classical methods in lesion detection and volume estimation \cite{Isensee2021,Commowick2018}. However, convolutional networks are limited in capturing long-range dependencies, which are important when lesions are small, irregular, and dispersed across the brain. Transformer-based architectures extend deep learning by incorporating self-attention mechanisms, which model global context more effectively. UNETR (U-Net with transformers as encoders) combines the strengths of U-Net’s hierarchical decoding with transformer-based feature extraction, while SegFormer uses lightweight hierarchical transformers to achieve efficient, resolution-robust segmentation \cite{Hatamizadeh2022,Xie2021}. These models are particularly well suited for 7T data, where both fine detail and broad spatial context are needed: lesions may be very small, but distinguishing them from artifacts requires global contextual cues. Comparing UNETR and SegFormer to standard convolutional models provides insight into whether transformer-based methods offer a meaningful advantage for ultra-high field MRI lesion segmentation.


Classical convolutional architectures such as U-Net and its ResNet-based variants remain the standard approach for medical image segmentation and form a strong baseline for MS lesion detection. Their strength lies in multi-scale feature extraction, but their reliance on local convolution can limit sensitivity to small, spatially dispersed lesions—an important consideration in 7T MRI. Transformer-based models mitigate this limitation by using global self-attention to integrate contextual information across the entire input, potentially improving discrimination between small true lesions and high-field artifacts.

The qualitative results help interpret these numeric differences in light of the underlying inductive biases. In Figure~\ref{fig:qualitative_main}, LST-LPA frequently labels bright cortical boundaries, choroid plexus, and septum pellucidum as lesions, illustrating its tendency toward over-segmentation in the presence of 7T-specific contrast features. LST-AI substantially reduces these false positives but occasionally underfills lesions or misses smaller foci compared with manual annotations, consistent with a convolutional model tuned to 3T contrast and artifact profiles. UNETR and SegFormer more closely follow the expert masks overall and, in some cases, recover small lesions that were not included in the original manual labels, consistent with the ability of transformer encoders to integrate information across the full field of view. At the same time, the artifact example in Figure~\ref{fig:artifact} shows that transformer-based models do not uniformly dominate: here, only SegFormer avoided labeling a wraparound artifact as lesion, whereas LST-LPA, LST-AI, and UNETR all generated false positives. These patterns suggest that architectures with stronger global context modeling can help distinguish small lesions from 7T-specific artifacts, but also that their performance remains bounded by label quality and the diversity of artifact patterns seen during training.

A key question for 7T imaging is whether high spatial resolution confers a practical advantage for automated lesion segmentation. By training UNETR and SegFormer at 0.5$\times$0.5$\times$0.5~mm\textsuperscript{3}, 1.0$\times$1.0$\times$1.0~mm\textsuperscript{3}, and 1.5$\times$1.5$\times$2.0~mm\textsuperscript{3}, and evaluating all outputs at the native 7T resolution, we observed a consistent degradation in performance as training resolution decreased. Dice and lesion-wise Dice declined at coarser resolutions, lesion-wise HD95 increased (including an infinite value for SegFormer at the coarsest resolution), and both TP counts and qualitative lesion coverage were reduced (Tables, Figure~\ref{fig:resolution}). These observations support the view that the fine spatial detail available at 7T is not merely a cosmetic difference but materially improves automated segmentation when models are trained directly in this high-resolution space.

Interpretation of the performance rankings in this study must consider the way reference segmentations were constructed. Initial lesion masks were generated using LST-LPA or LST-AI and then manually corrected by expert raters. LST-AI was used only in constructing the test-set reference masks and did not contribute to the label generation for the training set, where only LST-LPA outputs (followed by manual refinement) were used as the starting point. This design introduces the possibility that LST-AI benefits from partial circularity during evaluation, since its predictions on the test set may more closely resemble the pre-correction masks that guided manual editing. As a result, the overlap-based metrics reported for LST-AI are best interpreted as an optimistic upper bound for that tool under these specific labeling conditions, rather than as a fully independent benchmark. At the same time, the fact that transformer-based models trained solely on manually corrected labels—without any direct exposure to LST-AI outputs in training—approach LST-AI performance and sometimes offer qualitatively cleaner segmentations in challenging regions suggests that 7T-specific training can mitigate some of the domain shift faced by tools originally developed at 3T.

Another important consideration is that our evaluation focuses on 7T FLAIR acquired at a single institution using a specific scanner and protocol. Although the number of subjects (138 participants, 287 scans) is modest compared with some natural-image benchmarks, each full-brain 0.5$\times$0.5$\times$0.5~mm\textsuperscript{3} volume provides millions of labeled voxels and numerous lesions. In practice, this yields a large effective sample of lesion and non-lesion patches, which is sufficient for compact transformer architectures such as UNETR and SegFormer. Nonetheless, future work could investigate whether self-supervised or multi-site pretraining further improves generalization, particularly when extending to different scanners or 7T protocols. Finally, we did not systematically explore alternative loss functions, ensembling strategies, or pretraining schemes (for example, self-supervised pretraining on multi-contrast 7T data), all of which could further improve transformer-based performance.

In conclusion, the present study offers several contributions to the emerging literature on 7T MS imaging. First, it provides a systematic, lesion-wise evaluation of classical, CNN-based, and transformer-based segmentation tools on high-field FLAIR using metrics aligned with recent community standards. Second, it demonstrates that training directly on 7T data at native resolution yields measurable benefits in both quantitative metrics and qualitative lesion depiction, particularly for small lesions. Third, by making our 7T-trained models publicly available, we aim to lower the barrier for incorporating automated lesion analysis into 7T studies at centers that lack large annotated datasets. Future work should extend this framework to multi-center cohorts, incorporate gray matter and cortical lesions, and explore joint modeling of multi-contrast 7T sequences. More broadly, the present results suggest that transformer-based architectures, when trained on sufficient 7T data, are a promising complement to established tools and can help move the field toward more standardized and reproducible lesion quantification at ultra-high field strength.


\section*{Acknowledgments}
The support of NSF through grant 2020624 (OISE) is acknowledged (PI Fermuller). We wish to thank the technologists and MRI physicists at the Kennedy Krieger Institute and Johns Hopkins University for assistance with acquisition of the 7T MRI images. We also wish to thank Kerry Naunton and Christina Ecker for their work in patient recruitment and research coordination.

\section*{Funding}
Acquisition of MRI data in this study was funded by NIH R01NS104403 (Harrison, PI).

\section*{CRediT Authorship Contribution}
% ...






M. Maynord: Conceptualization, Software, Formal Analysis, Validation, Investigation, Methodology, Visualization, Writing – original draft, Writing - review and editing

M. Liu: Software, Formal Analysis, Validation, Investigation, Methodology

C. Ferm\"uller: Formal analysis, Funding acquisition, Methodology, Project administration, Resources, Supervision, Writing – review and editing

S. Choi: Data curation, Resources, Software, Writing - review and editing

Y. Zeng: Data curation, Investigation, Methodology, Writing - review and editing

S. Dahal: Data curation, Investigation, Methodology, Writing - review and editing

D.M. Harrison: Conceptualization, Data curation, Funding acquisition, Investigation, Methodology, Project administration, Resources, Supervision, Validation, Writing – original draft, Writing – review and editing

\section*{Data and Code Availability}


Model weights for UNETR and SegFormer at all three resolutions are available here: \href{https://github.com/maynord/7T-MS-lesion-segmentation}{https://github.com/maynord/7T-MS-lesion-segmentation}


The third-party tools used as baselines in this work (LST-LPA and LST-AI) are distributed by their original authors and can be obtained from the corresponding software repositories as described in the cited publications.

At present, the underlying individual-participant MRI data cannot be deposited in a public repository because of institutional review board and participant privacy constraints. De-identified derived outputs at the level of lesion masks and aggregate metrics may be shared for qualified research collaborations upon reasonable request to the corresponding author, contingent on appropriate data-use agreements and local regulatory approval.

\section*{Use of AI and AI-assisted Technologies}
During the preparation of this work the authors used ChatGPT in order to aid manuscript drafting. After using this tool, the authors reviewed and edited the content as needed and take full responsibility for the content of the published article.

\section*{Declaration of Competing Interest}
Dr. Harrison receives research funding from TG Therapeutics and Roche Genentech. Dr. Harrison has received consulting fees from TG Therapeutics and royalties from Up To Date, Inc.


\bibliographystyle{elsarticle-num-names}
\bibliography{2025_09_24_ms_lesion_references}




\end{document}